\documentclass[procedia]{easychair}

\usepackage{doc}
\usepackage{makeidx}

\usepackage{tabularx}
\usepackage{graphicx}
\newenvironment{myfont}{\fontfamily{cmr}\selectfont}{}

\def\procediaConference{Biologically Inspired Cognitive Architectures 2015 (BICA 2015)}

\title{Using a Distributional Semantic Vector Space with a Knowledge Base for Reasoning in Uncertain Conditions}

\titlerunning{Reasoning in Uncertain Conditions}

\author{
    Douglas Summers-Stay
\and
   Clare Voss
\and
   Taylor Cassidy
}

\institute{
 U.S. Army Research Laboratory\\
  \email{douglas.a.summers-stay.civ@mail.mil}
 }

\authorrunning{Summers-Stay, Voss and Cassidy}

\begin{document}

\maketitle

\begin{abstract}
The inherent inflexibility and incompleteness of commonsense knowledge bases (KB) has limited their usefulness. We describe a system called Displacer for performing KB queries extended with the analogical capabilities of the word2vec distributional semantic vector space (DSVS). This allows the system to answer queries with information which was not contained in the original KB in any form. By performing analogous queries on semantically related terms and mapping their answers back into the context of the original query using displacement vectors, we are able to give approximate answers to many questions which, if posed to the KB alone, would return no results.

We also show how the hand-curated knowledge in a KB can be used to increase the accuracy of a DSVS in solving analogy problems. In these ways, a KB and a DSVS can make up for each other's weaknesses.
\end{abstract}


%
%

\section{Introduction}

A knowledge base (KB) is a database of facts about the world, together with an inference engine to perform deductive reasoning on those facts. Most of the facts in a KB are expressed using first-order predicates, creating links between two objects in the database that expresses their relationship, such as (capitalCity Paris France). The brain organizes long term, explicit, semantic memory in a way that is very different from a knowledge base. We seem to be able to call up a concept from memory
\begin{itemize}
\item by thinking of concepts whose meaning is similar.
\item by completing an analogy.
\item based on attributes, including attributes we didn't previously realize characterized that concept.
\item by seeing an example of the concept.
\item by deductive reasoning based on related concepts.
\item by calling to mind an image and reasoning about details of the image.
\end{itemize}

Suppose I am looking for a specific animal-- a cardinal, for example.  If I had a set of facts in my memory like ``you can find things where they live" and ``all birds live in the woods" and ``cardinals are a type of bird" then I could reason using deductive logic that I should look for cardinals in the woods. But I don't actually have many facts which are universally true in my memory, and deductive reasoning relies heavily on universal statements. What I have instead are many specific pieces of knowledge, which are useful only for analogical reasoning: I have often heard bluejays in trees, and cardinals look a little like bluejays (except red instead of blue and black) and there are a lot of trees in the woods. I know that hummingbirds live in tiny nests in small trees and eagles build enormous eyries in tall trees (which is nearly the same thing as a nest), and cardinals are somewhere in between, so (reasoning by a kind of analogy) I am looking for a nest a few times the size of a cardinal in an average-sized tree. Maybe I could find them by looking for what they like to eat, but I'm not sure exactly what cardinals eat. I'm nearly certain sparrows like to eat small seeds, and I'm pretty sure birds eat little bugs. Maybe larger birds eat larger bugs? But blue whales eat krill, which is very tiny, so not always, though if a bigger bird eats smaller bugs it must eat a lot more of them... and so on.  It is a mess of biased reasoning, stereotypes, prejudice, ignorance, and unfounded analogy; but I come to a sensible conclusion that if not exactly correct is workable, at least. When it is a problem we know very little about, such reasoning takes a little time and effort, but many times a day we solve similar problems nearly instantly and without conscious thought. 

If we try to imagine doing this kind of thing with a knowledge base, it looks impossible. How could we ever hope to encode all the facts we know about these subjects, and to what degree they are certain, and in what ways they are related? How could we hope to build a reasoning system that can handle the kind of fuzzy, approximate, analogical, incorrect reasoning I am describing, and come up with any kind of useful results? The system described in this paper is an attempt to solve such problems.

Since the 1970s, researchers have been attempting to build large knowledge bases that contain much of the common sense knowledge that we make use of whenever we interpret a sentence or answer a question. The results of this effort have been disappointing, due to two main difficulties. The first is that a KB is inflexible: if a query is not entered using exactly the right terms and predicates that the creators of the KB envisioned, it gives no answer at all. The second problem is that all KBs are incomplete: because all facts need to be entered by hand, even decades of work have left enormous gaps in what KBs contain about the world.

In the last few years, a new approach has appeared to encoding semantic relations that implicitly contains a great deal of the same kind of world knowledge we would like to have in a KB. Using a large corpus of natural language text, a distributional semantic vector space (DSVS) can be automatically created. A DSVS represents natural language words and phrases as high-dimensional vectors. Vectors which are nearby\footnote{Nearness in a DSVS is typically measured using a cosine metric or a Euclidean metric.} in this space are semantically similar. A remarkable property of these vectors is that one can perform a kind of analogical reasoning by simply doing arithmetic with these vectors: subtract the vector for the word `man' from the vector for `king' and add the vector for `woman' and one gets a vector very close to the vector for the term `queen.' This ability to automatically find analogies has enormous potential, and research into the area has exploded since the word2vec DSVS was introduced in \cite{Mikolov2013}. The challenge in using a DSVS is finding ways to handle approximate and implicit knowledge.

Suppose we want the answer to a query using a term that has no facts asserted regarding it in the KB. A DSVS allows us to find related terms which are in the database, and find the answers to those related queries. This is very useful at overcoming the KB \emph{inflexibility} problem. However, the word2vec DSVS allows us to go one step farther, and address the KB \emph{incompleteness} problem as well. We can make use of these related terms and their associated query results to set up an analogy problem with the original unknown term so that its solution (obtained from the DSVS) is the answer to the original query. Both the KB and DSVS have explicit maps to natural language, which allows us to connect concepts in the KB to concepts in the DSVS and back by way of the natural language term they are both associated with. We have built a system called Displacer that connects a KB and DSVS in this way, and demonstrated the increases in coverage and accuracy that the system produces on a range of tasks. 

\section{Biological Inspiration}
In 1984, Geoff Hinton \cite{hinton} outlined ways in which distributed representations, characteristic of a DSVS, were more biologically plausible than the local representations used in KBs. He pointed out that both the strengths of human memory (content-addressible, generalization, analogy making) and its weaknesses (the difficulty of remembering an arbitrary string of concepts) are similar to those of distributed representations.

Brain-imaging studies have likewise suggested that concepts are represented in the brain as distributed networks of neural activation \cite{rissman}. A DSVS can be interpreted as a neurally plausible model of how memories could be encoded. \cite{blouw}. In particular, the analogical-reasoning capability of a DSVS can be understood as an example of the relational priming model of analogy making outlined in \cite{leech}. There is evidence that object categories are represented as a continuous semanatic space across the surface of the brain. \cite{huth} The brain's slow operating speed and massive paralellism (compared to a CPU) also hint that whatever operations are being performed must be very short, simple programs operating on large vectors, more characteristic of A DSVS than a KB. Together, all of these clues suggest that whatever representation of concepts is used in the brain, it is much more similar to a DSVS than a KB.

The system described in this paper is an attempt to take advantage of the benefits that these more neurologically plausible representations of concepts allow. Its hybrid nature is probably merely an intermediate step towards an integrated model in which all the operations in deductive reasoning are also carried out on the vectors themselves. In the meantime, however, the KB, while not itself biologically inspired or plausible, is convenient for applying mutiple Horn rules and keeping track of the steps of the reasoning process in a way that is not yet convenient in a DSVS.

\section{Background}

Here we describe how a KB and a DSVS are used, and connect this effort with previous research into combining the two. We have not yet tackled the problem of parsing natural language queries into KB queries, though that would be an important component of a finished system.

\subsection{Answering Queries with the ResearchCyc KB}

Queries in ResearchCyc are formulated in a lisp-like language called CycL. A typical query is \begin{myfont}(capitalCity ?X France)\end{myfont}. The predicate, \begin{myfont}capitalCity\end{myfont}, is the first term in the triple. The second term in this case is a variable, represented as \begin{myfont}?X\end{myfont}. The answers to the query are the expressions in CycL which substituted for ?X make a true statement, as deduced from the knowledge and rules in the KB.

The ResearchCyc KB contains natural language paraphrases for most concepts in the database, and has the ability to convert noun phrases into the corresponding Cyc representation. For example, the noun phrase ``rich ruler'' returns the following possible Cyc interpretations:

\fbox{
 \begin{minipage}{.95\textwidth}
\begin{myfont}word2cyc\end{myfont}

Cyc command:

\begin{myfont}
 (ps-get-cycls-for-phrase ``rich ruler'')\end{myfont}

Results:

\begin{myfont}
(SubcollectionOfWithRelationToFn \textbf{Leader} personalWealthOwned \textbf{GreatWealth})\end{myfont}\footnote{i.e. ``wealthy leader''}

\begin{myfont}
(SubcollectionOfWithRelationToFn \textbf{Leader} affiliatedWith Mark\textbf{Rich})

(SubcollectionOfWithRelationToFn \textbf{Leader} affiliatedWith Adrienne\textbf{Rich})

(SubcollectionOfWithRelationToFn \textbf{Leader} affiliatedWith \textbf{Richard}Cronin-Musician)

(CollectionIntersection2Fn \textbf{Leader} (SubcollectionOfWithRelationToFn HomoSapiens lastName ``\textbf{rich}''))

(SubcollectionOfWithRelationToTypeFn \textbf{Leader} affiliatedWith (SubcollectionOfWithRelationToFn HomoSapiens lastName ``\textbf{rich}''))\end{myfont}\footnote{i.e. ``leader named (or affiliated with someone named) Rich''}
\end{minipage}
}

The possible interpretations of ``rich'' in terms of food (containing sugar and cream) and ``ruler'' as a tool (a shorter yardstick) are ruled out by the KB due to their incompatibility with each other:``rich'' in that sense can only describe food, and ``ruler'' in that sense cannot own wealth.

In addition, Cyc has a limited ability to generate natural language paraphrases for the statements it generates in response to queries. For example, feeding the first Cyc result for ``rich ruler'' above back into the KB gives the following (mildly incorrect and stilted English) paraphrase:

\fbox{
 \begin{minipage}{.95\textwidth}
\begin{myfont}cyc2word\end{myfont}

Cyc command: 

\begin{myfont}(generate-text-w/sentential-force '(\#\$SubcollectionOfWithRelationToFn \#\$Leader \#\$personalWealthOwned \#\$GreatWealth) )\end{myfont}

Result:

\begin{myfont}``leader whose personal wealth is: wealth''\end{myfont}
\end{minipage}
}

These two functions (which we call \begin{myfont}word2cyc\end{myfont} and \begin{myfont}cyc2word\end{myfont} to be consistent with the name of the function \begin{myfont}word2vec\end{myfont}), allow the Displacer system we've constructed to take a natural language term or phase, include it in a query, and map the answer to the query back into natural language (see Fig. 1). While ResearchCyc has a large vocabulary, many terms do not map to any expression in the KB. In contrast, the DSVS contains essentially all commonly used English words. We additionally used synonym and part-of-speech information from wordnet to augment the KB for some tasks.

\begin{figure}[h]
\includegraphics[width = 0.95\textwidth, clip = true, trim = 0cm 18.5cm 7cm 1cm, page=3]{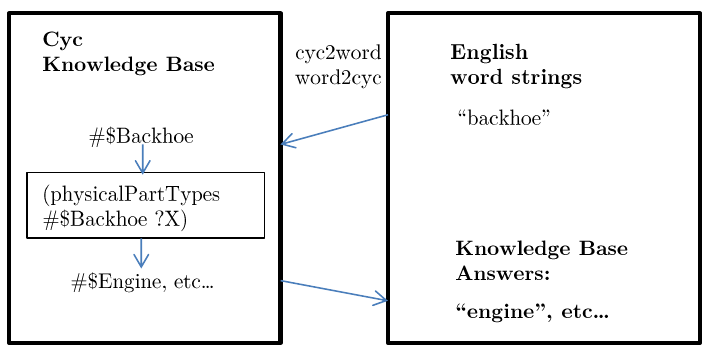}
\caption{Answering queries with the Cyc KB}
\end{figure}

Using the ResearchCyc KB to support web queries is explored in \cite{Conesa2007}. They write, ``Although problematic, it would be worthwhile to use ResearchCyc to support web queries. The reason is that, as far as we know, it is the only ontology that contains linguistic, semantic and factual knowledge.'' This is a major reason we chose to use ResearchCyc as the KB in these experiments as well.

\subsection{Solving Analogies with the word2vec DSVS}
Proportional analogies were introduced by Aristotle. In the 1980s and early 90s, researchers attempted to solve analogy problems using KBs. \cite{Gentner1983} suggested using the structure of concepts to find analogies, rather than the more obvious matching of relations. Unfortunately, as \cite{Hofstadter1994} pointed out, such structure is heavily dependent on the preconceptions inherent in the design of the knowledge base. The use of distributional semantic information was a breakthrough, allowing the system of \cite{Turney2005} to score at human level on multiple-choice SAT analogy questions. The word2vec DSVS \cite{Mikolov2013} uses an efficient method to build the vector space, allowing it to be created with hundreds of billions of words of training data. Except for this capability of using a larger training set and some well-chosen parameters, however, word2vec is essentially optimizing the same objective as the latent semantic analysis (LSA) approach to building semantic vectors.

The word2vec DSVS represents English words and phrases as points in a high-dimensional space. These points have the property that the relationship between two words is to some extent encoded by the vector between them. For example, the vector for ``king'' with the vector for ``man'' subtracted from it is a new vector encoding something about the difference between an ordinary man and a man who is a king. This can be thought of as a predicate relation we could call ``royal'' that maps ``man'' to ``king:''

\begin{myfont}
(royal man ?X)$\Rightarrow X = $ king\end{myfont}

When applied to another term such as ``woman,'' this returns a vector near to the vector for ``queen:''

\begin{myfont}
(royal woman ?X)$\Rightarrow X \approx $ queen\end{myfont}\footnote{The basic reason why word2vec is able to solve these kinds of analogies is explained clearly in \cite{Arora2015}.}

\begin{figure}[h]
\includegraphics[width = 0.95\textwidth, clip = true, trim = 0cm 16.5cm 6cm 1.3cm, page=1]{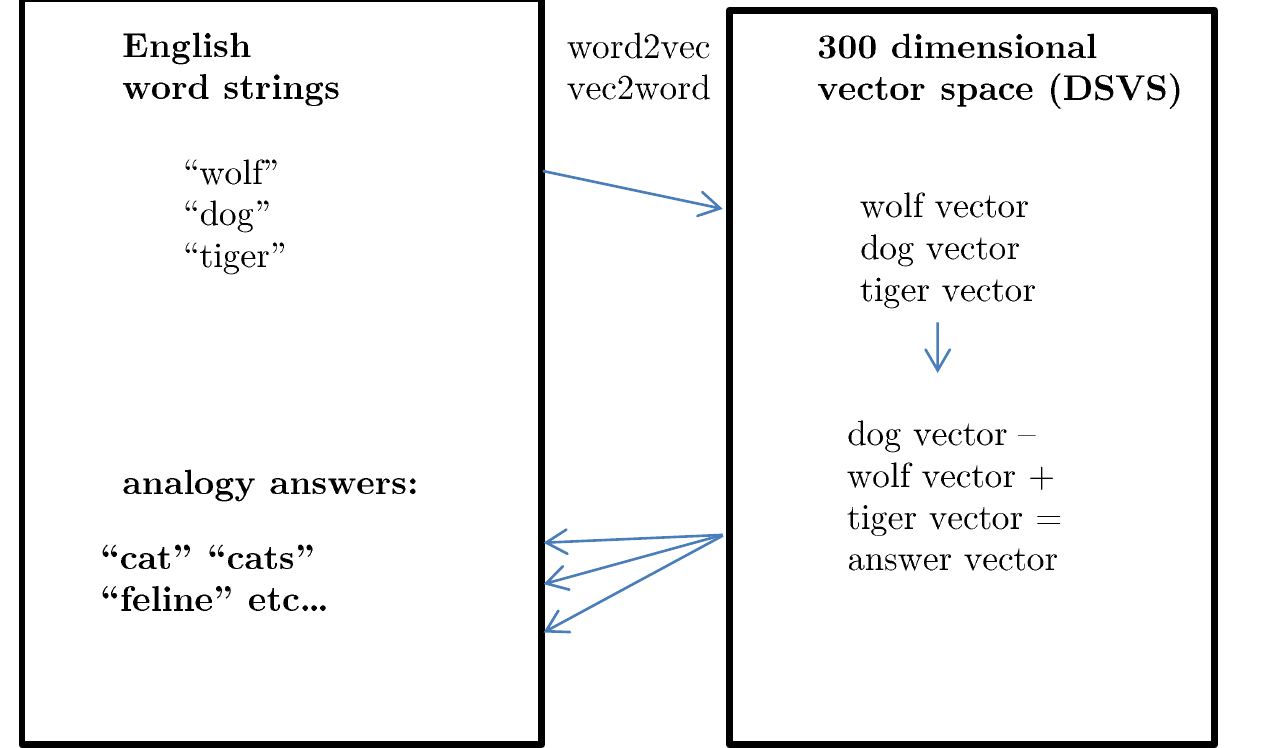}
\caption{Solving analogies with the word2vec DSVS}
\end{figure}

 There are a few differences between predicates in a knowledge base and these relation vectors in the DSVS. First, each relation vector is only approximate. Some of them, such as the ones encoding royalty, plurality, or capital cities, are close approximations that are valid for most terms on which they are defined. Others, such as the ones encoding hypernyms, vary a lot from one term to another \cite{Rei2014}.

A second difference is that a predicate can return zero, one, or many results, while a relation vector is limited to a single result. In some cases returning multiple near neighbors to the resultant vector gives useful results, but for other types of predicates the results have a different structure. For example, the hypernyms of a given term are not clustered together in semantic space but instead form a tree-like structure with the term at a leaf of the tree.

For these reasons, one cannot simply define a single vector, or even a tensor (a mapping from the full vector space into the same vector space) that represents a predicate correctly in many cases. However, assuming that each individual relation vector (representing a mapping between a term and one of its query results) is applicable to nearby semantic vectors, we can use relations that we know to be true to apply to nearby search terms to find their analogous results. This is making the assumption that predicate relation vectors form a kind of manifold, where the closer to the original term in semantic space, the more likely the relation will be accurate. In the case of one-to-many predicates, we will need to find a unique vector for each of the many search results and apply each to the new search term to get the corresponding result.

We used a version of the word2vec DSVS trained on over 100 billion words of text from Google News. This contains the most common million words and short phrases (up to three words) from the dataset. There was little preprocessing done on this data so words exist in both capitalized and commonly misspelled forms. (Such uses are usually very close semantically.) Each term or phrase is represented by a 300 dimensional vector.

The functions \begin{myfont}word2vec\end{myfont} and \begin{myfont}vec2word\end{myfont} map natural language words and phrases into the DSVS. If the phrase appears verbatim within the DSVS,  \begin{myfont}word2vec\end{myfont} returns that vector. If not, the vector returned is the average of the vectors for each of the words in the phrase individually (ignoring articles and the word ``of''). This average has been shown to be semantically near to the relevant concept in the DSVS in many cases.\footnote{The sum and the average are identical when semantic distance is measured using a cosine distance, as it is in many of these papers.} The function \begin{myfont}vec2word\end{myfont} returns the most similar words and phrases to a given vector in the DSVS. We use approximate nearest neighbors \footnote{A Matlab wrapper by Shai Bagon for a k-d tree ANN library by David Mount and Sunil Arya.} to quickly find similar vectors to a given vector and return the associated word or phrase.

It is important to note that the DSVS may not have learned all the analogies which we would potentially like to make use of. Finding analogies between terms and their antonyms, for example, proved to be surprisingly difficult for word2vec.

\subsection{Using a KB with a DSVS}
In \cite{Mikolov2013} the creators of word2vec write, ``Our ongoing work shows that the word vectors can be successfully applied to automatic extension of facts in Knowledge Bases, and also for verification of correctness of existing facts.'' They see the DSVS as being used as a resource to add assertions into knowledge bases and to verify existing assertions. The work described in this paper uses the DSVS for a similar purpose, but instead of asserting the facts directly, obtains them from the DSVS as needed. The disadvantage of this on-the-fly approach is that the assertions can't be used as intermediate steps in the reasoning process. However, any fact whose only source is a DSVS is inherently uncertain, and this makes adding them directly to the KB problematic. Our approach uses hand-verified assertions throughout the reasoning process, and only at the end brings in analogical reasoning through the DSVS. 

There has been some research into using semantic similarity to perform queries not just on the input terms but on similar terms as well. This can take place throughout the reasoning process and on the final results. A good survey of query expansion is \cite{Carpineto2012}. The idea of using a DSVS to extend the capability of a KB as a query-KB semantic matching technique is explored in \cite{Freitas2014distributional}. The approach described in this paper is similar, but the ability to make use of the analogy-forming properties of the DSVS (to map the results back into the context of the original search terms) allows it to answer many queries which would be impossible in these other systems.

There has been some exploration of how various relations are encoded into the DSVS. \cite{Rei2014} for example, explores how hyponyms swarm around a term. The possibility has also been explored of reshaping a DSVS according to verified facts in \cite{Faruqui2014}. One problem with doing this is that other relations, not explicitly included in such reshaping, may be distorted and so no longer have the analogical properties they had in the original DSVS. \cite{Wang2014} also explores putting a knowledge graph (that is, a KB) into the space of a DSVS.

Ultimately, maintaining a KB and a DSVS separately does not seem to be the most elegant solution to the problem, however. Methods that create or modify a vector space from a KB directly allow the two to be combined in a single representation which could be used directly for both deductive and analogical reasoning. AnalogySpace, built from the large KB ConceptNet\cite{Speer}, is one such example. Knowledge graph embedding seems promising, especially when combined with knowledge extraction efforts such as KNEXT\cite{schubert} and NELL\cite{carlson}. At the moment, however,  a KB still provides certain features these systems do not, such as making explicit the chain of reasoning to reach a conclusion. 

\section{Experiments with Answering Queries}

 The first experiment (3.1) demonstrates how semantic similarity can be used to estimate which of two answers to a query is more likely to be correct, based on the results of the query on similar terms. The second experiment (3.2) is an attempt to estimate the likelihood of finding the best answer to a query within the first few results returned by Displacer. The third experiment (3.3) shows how Displacer can be used on queries with many correct answers. The last two experiments (4.1 and 4.2) use the components of Displacer in a different arrangement to show how the KB can be used to enhance the performance of the DSVS in solving analogies, rather than answering queries. The combination of a KB and DSVS is shown to be better at the tasks designed for either one.

\subsection{Experiment: Estimating the gender of given names}

The purpose of this experiment was to test the ability of Displacer to correctly answer a query on an unknown term when there are two likely possibilities for the result in the knowledge base. We used data from the 1990 U.S. Census Bureau on the 800 most popular names for men and women in the U.S (which do not appear on the list for the other gender).  We queried the system on the gender of the individual with each name. The KB contains gender and name information for some celebrities and historical individuals. Of the 1600 names, only 146 were recognized as belonging to some individual about which the KB had gender information. For the rest, it estimated the answer based on the responses to similar terms. Responses where more than 50\% of names received the label \begin{myfont}male\end{myfont} were classified as male names, and similarly for female names.

\subsubsection{Approach}

Displacer was used as follows for this experiment: 

\begin{enumerate}
\item Begin with a list of the 800 most popular uniquely male and female names from U.S. census data. For each name, do the following:
\item Map the name from an English word string to a vector using word2vec.
\item Find the nearest neighbors of the vector in the DSVS.
\item Map the nearest neighbors from vectors to English words using vec2word.
\item Map the English words to Cyc constants using word2cyc.
\item Search Cyc for famous individuals with the same given name, and return the gender of those famous individuals.
\item Map the resulting Cyc expressions into English words (`male' and `female') using cyc2word.
\item Map the results from the previous step to vectors using word2vec.
\item Average the vectors and compare the distance from the average vector to the vectors for `male' and`female.'
\end{enumerate}

\subsubsection{Results}

\begin{table}[h]
\caption{Estimating gender of given names}
\label{sample-table}
\begin{center}
\begin{tabularx}{0.95\textwidth}{ |l|X|X| }
\hline
 &labeled male &labeled female\\
\hline
KB: of 800 men &117 &3\\
KB: of 800 women &6 &37 \\
Displacer: of 800 men &723 &77\\
Displacer: of 800 women &82 &718\\

\hline
\end{tabularx}
\end{center}
\end{table}

\subsubsection{Discussion}

The methodology here could perhaps be improved. The names in the database come from all over the world (though with a heavy western bias), while in the test set they only come from the United States. Regardless, the system is clearly making correct inferences in most cases. The underlying reason for this simple. Female names are often used in sentences together with the word `she' in similar contexts, leading them to be mapped close together in the DSVS. The vector leading from any one of these names to the vector for `Female' is close to identical (and similarly for male names and the vector for `Male'). 

\subsection{Experiment: Estimating probability of finding single correct answers}

In the previous experiment, there were only two possible answers, and it was only necessary to discriminate between them. For other predicates such as \begin{myfont}capitalCity\end{myfont} or \begin{myfont}animalTypeMakesSoundType\end{myfont} querying on any term gives a unique result. Displacer is able to take all of these unique results and create a combined estimate for queries on terms not in the KB. In the word2vec DSVS, the vector from term A (e.g. France) to its result B (Paris) can be applied to term A' (England) to estimate its result B' (London). By averaging several of these vectors, we can reduce noise and obtain a high-quality estimate.

Choosing A as near as possible to A' is expected to give the best results, since the contexts will be most similar and the vector between A and B most nearly parallel to the vector between A' and B'. However, there will inevitably be noise in the context, and averaging more vectors together will tend to reduce this noise. Empirically, these two competing tendencies tend to give a minimum between 3 and 5 neighbors averaged, and follow a general shape of a dip followed by a rise that levels off, as seen in Fig. 3.
%

\subsubsection{Approach}

In this experiment, we chose predicates for which each query returns only a single result, and most of those results were unique. 
The steps that Displacer followed in this experiment were similar to those in the previous experiment, but the results were mapped back using the displacement vectors in the final steps. As a concrete example, the procedure below was followed for the first query.
\begin{enumerate}
\item Query the KB to obtain a list of countries.\footnote{In this experiment, we used a leave-one-out methodology since the goal was to estimate how often using the full KB would give the right answer.} For each country, do the following:
\item Map the name of the country from an English word string to a vector using word2vec.
\item Find the nearest neighbors of the vector in the DSVS.
\item Map the nearest neighbors from vectors to English words using vec2word.
\item Map the English words to Cyc expressions using word2cyc.
\item Search Cyc for the capitalCity of the constants in the previous step. If the expression is not a country, it will have no capital city and be ignored.
\item Map the resulting Cyc expressions into English words (the names of capital cities) using cyc2word.
\item Map the results from the previous step to vectors using word2vec.
\item Using the similar country vector from step 3, the similar capital vector from step 9, and the original country vector from step 2, calculate 

(estimated capital vector) = (similar capital vector) - (similar country vector) + (original country vector)

\item average the estimated capital vectors from all near neighbors.
\item map the average estimated capital vector from the previous step to the closest English words using vec2word.
\end{enumerate}

\subsubsection{Results}

\begin{table}[h]
\small
\caption{Probability of the $n^{th}$ closest word to the average estimated vector being correct for a particular query}
\label{sample-table}

\tabcolsep=0.11cm
\def\arraystretch{1.5}
\begin{tabularx}{0.63\textwidth}{ |l|l|l|l|l| }
\hline
query &1st &2nd &3rd &4th\\\hline
(and (capitalCity ?Y ?X) & & & & \\ (isa ?Y Country)) &.30 	&.19 	&.03 	&.01\\\hline
(and (monetaryUnitIssuedBy ?X ?Y) & & & & \\(isa ?Y Country) ) &.25	&.05	&.12	&.04\\\hline
(agentTypeUsesArtifactType ?X ?Y) &.13	&.02	&.05	&.02\\\hline
(animalTypeMakesSoundType ?X ?Y) &.05	&.05	&.15	&.05\\\hline
(teamRepresentsPolity ?X ?Y)&.78 &.21 &.04 &.00\\\hline
(cityInState ?X ?Y)&.44	&.21 &.06 &.03\\\hline
(and (topicOfIndividual ?X ?Y) & & & &  \\(isa ?Y ScientificFieldOfStudy))&.12	&.05	&.30	&.05\\\hline
(countryOfCity ?Y ?X) &.52	&.17	&.03	&.01\\\hline
(provenanceOfMediaSeriesProduct ?X ?Y)&.46 &.17	&.07	&.02\\\hline
\end{tabularx}
\end{table}

\subsubsection{Discussion}

These probabilities apply to other query terms which are in the DSVS but not the KB. The query \begin{myfont}(capitalCity Germany ?X)\end{myfont} returns no results from the KB, because Germany was not yet a reunited country when the information was entered. However, because most other capital cities and their corresponding countries are included in the KB, the analogy between many of those countries and their capitals and between Germany and Berlin is very well supported, and the correct answer ,\begin{myfont} Berlin\end{myfont}, has a lower sum of distances than other nearby terms, such as \begin{myfont}Frankfurt\end{myfont}.

This probability estimate is not perfect. It may be the case, for example, that certain facts not asserted in the KB are less frequently encountered facts, in which case the DSVS is going to have a poorer estimate of those facts as well. The process also requires knowing what general type of answer is acceptable for the given predicate and term, in order to take the sample from which to build the estimate.\footnote{For instance, the predicate \begin{myfont}capitalCity\end{myfont} was only applied to countries in the above example.} In some cases, the restrictions the KB places on an argument are enough. When this is insufficient, either direct user input or using the KB to find the nearest common generalization of some user-provided examples would be required.

Each of the queries above has only one correct answer for each input. Queries with multiple correct answers would be more difficult to assign a probability to in a reasonable way. \footnote{When capital cities and currencies show up again in the analogy experiment, we are not using information about many capital cities or many currencies, but working from only a single example to either find the relevant predicate or calculate the analogy directly. In the rare case where the predicate can be determined from the first two terms but is not defined on the third term, using the method outlined here could have given somewhat better results in that experiment.}

\subsection{Experiment: Using a DSVS to support a KB in answering queries with many results}

In both experiments above, each query returned only a single result. In queries asking for the parts of objects, for example, there are many correct results for each query. In such cases Displacer can still be used, but the probability of a result being correct cannot be calculated in the same way. Instead, we use k-means to cluster the results, and find the English word corresponding to the mean of each cluster. 

\subsubsection{Approach}

\begin{enumerate}
  \item Using Wordnet, we created a list of 346 machines (hyponyms of the word \begin{myfont}machine\end{myfont}). For each machine on the list (e.g. backhoe), follow the steps below (also in Fig. 4):
  \item Map the term \begin{myfont}backhoe\end{myfont} to a vector using word2vec.
  \item Find the nearest neighbors of the vector in the DSVS.
\item Map the nearest neighbors from vectors to Englsih words using vec2word.  The ten most similar terms in the DSVS are:\begin{myfont}
backhoe, excavator, trackhoe, bulldozer, payloader, bucket loader, Bobcat loader, skid loader, dump truck, backhoe operator\end{myfont}. 
  \item Map the English words to Cyc expressions using word2cyc. The following terms are found in the KB and have some physical parts defined: 
\begin{myfont}
\#\$Backhoe, \#\$Bulldozer, \#\$DumpTruck, \#\$Tractor, \#\$WreckerTruck, \#\$Shovel, \#\$Forklift, \#\$Truck 
\end{myfont}
  \item Search cyc for the parts of the machines in the previous step using the query 

\begin{myfont}
(and (physicalPartTypes backhoe ?X) (genls ?X \#\$SolidTangibleArtifact)).
\end{myfont} 

 For example the query run using the first sense of \begin{myfont}shovel\end{myfont} is \begin{myfont}(physicalPartTypes \#\$Shovel ?X)\end{myfont} This returns the following parts of a hand shovel: \begin{myfont}\#\$Handle, \#\$ShovelBlade\end{myfont}. Similarly, the physical parts of a backhoe, a tractor, and so forth are found. Note that for some of these terms, the physical parts are derived using a chain of reasoning rather than asserted directly. For example, a \begin{myfont}\#\$WreckerTruck\end{myfont} is a type of \begin{myfont}\#\$RoadVehicle-InternalCombustionEngine\end{myfont}, so parts like \begin{myfont}\#\$Piston\end{myfont} and \begin{myfont}\#\$GasCap\end{myfont} which are defined for the larger category are inherited by the particular machine. Because this part of the KB is very incomplete, \begin{myfont}\#\$Backhoe\end{myfont} was not defined as a road vehicle, though \begin{myfont}\#\$Bulldozer\end{myfont} was.
  \item Map the resulting Cyc expressions into English words (the names of parts) using cyc2word. For example, the KB constant \begin{myfont}\#\$Shovel\end{myfont} returns \begin{myfont}shovel\end{myfont}.
  \item Map the results from the previous step to vectors using word2vec.
  \item Calculate the displaced vector for each part. This replaces the concept of \begin{myfont}shovel blade\end{myfont} in the context of a hand shovel with the analogous term in the context of a backhoe. It is how the DSVS completes the analogy \begin{myfont}shovel blade : shovel :: ??? : backhoe\end{myfont}.  For the vectors representing \begin{myfont}shovel blade – shovel + backhoe\end{myfont} the most similar terms are \begin{myfont}bucket loader\end{myfont} and \begin{myfont}excavator\end{myfont}. Other terms similar to the vector for \begin{myfont}shovel blade\end{myfont} include blade servers (a type of computer), retractable blades, and so forth. By telling it we are interested in near-synonyms of \begin{myfont}shovel blade\end{myfont} that are more similar to \begin{myfont}backhoe\end{myfont} than \begin{myfont}shovel\end{myfont} we improve the quality of the results. 
\item Use k-means to cluster the part vectors for this machine.
\item Map the mean of each cluster to an English word using vec2word. The list of most likely possible parts of a backhoe is:

\begin{myfont}
backhoe, clutch, engine compartment, steering mechanism, auto body, automotive suspension,  strut, brake, bulldozer, chassis, control device, drive train, drivers seat, fuel tank, internal combustion engine, transmission, windshield, land transportation wheel, tractor, wheel and axle, blade, grinder, submarine sandwich, handle, shovel, bucket, curved handle, auto part, forklift, horizontal stabilizer, tail, trailer, vertical stabilizer, lawn mower, cab, car battery, car engine, engine starter, exhaust system, four cycle engine, fuel gauge, gas cap, gas tank, muffler, odometer, oil filter, speed indicator, truck tire, truck wheel\end{myfont}.
\end{enumerate}

There are several incorrect results (\begin{myfont}submarine sandwich\end{myfont}, for example) mixed in with the correct results. Depending on the amount of coverage the KB has in a particular area, the quality of the results can vary. For applications with some room for error or that can interact with a human to check results, having a list of possibilities with noise mixed in can be more useful than having no results at all. There is also the possibility of using Displacer itself to further narrow down the results, by checking, for example, the size or material of the purported parts.

\begin{figure}[h]
\includegraphics[width = 0.95\textwidth, clip = true, trim = 0cm 7cm 0cm 1cm, page=2]{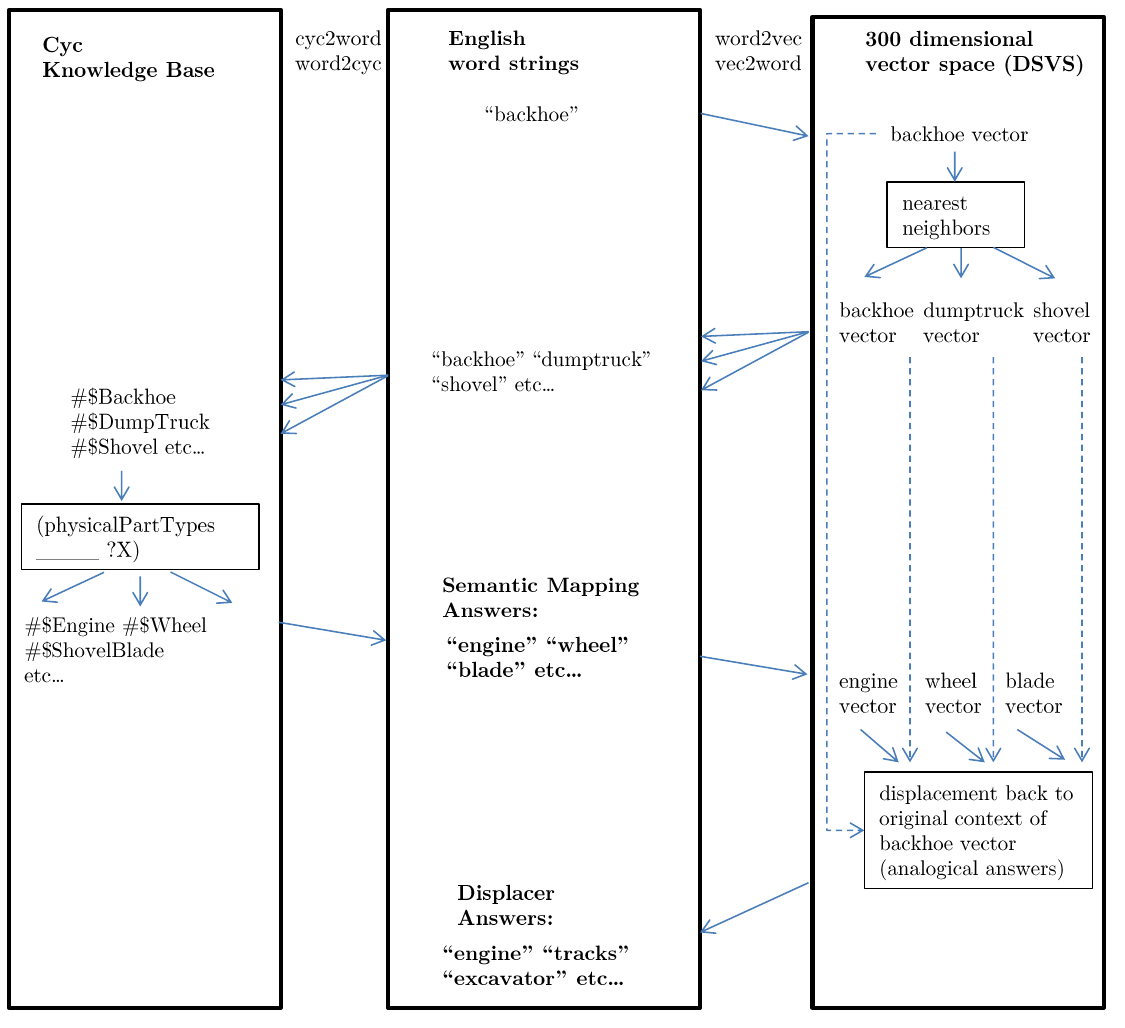}
\caption{Answering queries with Displacer}
\end{figure}

\subsubsection{Results}

 The KB was able to find at least one possible interpretation of 199 of these, missing 147. The DSVS (and the overall system) was able to find an interpretation of 294 of these, missing only 52. (These tended to be older terms that appear in dictionaries but rarely in newspaper articles such as \begin{myfont}gangsaw\end{myfont}, \begin{myfont}pavoir\end{myfont}, and \begin{myfont}triphammer\end{myfont}.) For the 294 machines, we asked Displacer for physical parts of these machines. For a fair comparison, we also included an incomplete versions of our full system. First we used the system in a query-KB semantic matching mode, similar to the system described in \cite{freitas2014natural}. This performs a query for similar terms, but does not perform the final displacement step that maps answers back into the original context using the analogy capability of word2vec.
Results are summarized in the table below.

\begin{table}[h]
\caption{Correct parts of machines found}
\label{sample-table}
\centering
\begin{tabularx}{0.585\textwidth}{ |l|l|l| }
\hline
&recognized & \% correct of top three \\
& & results for each\\\hline
ResearchCyc &199 / 346 & -- \\\hline
ResearchCyc &294 / 346 &39\% \\
with semantic & & \\
matching & & \\\hline
Displacer &294 / 346 &60\% \\

\hline
\end{tabularx}
\end{table}

\subsubsection{Discussion}

This shows that Displacer returns more correct answers than the KB alone or either of the systems that don't make use of the analogical capability of word2vec. In all the experiments above, we are applying a single predicate to a single term, but because Displacer builds on the full expressive power of the KB, it can be used with much more complex queries involving logical expressions (`and', `or', `not', `there exists', and so forth).

\section{Experiments with Solving Analogies}

The previous experiments showed how a DSVS can be used to extend the coverage and precision of a KB. In this experiment, we tested how a KB can be used to improve the analogical reasoning capability of a DSVS. In some cases, the relationship between the first two terms in a proportional analogy problem corresponds exactly to a predicate in a knowledge base. For example, in the analogy
\begin{myfont}
Paris : France :: London : X
\end{myfont}
the predicate \begin{myfont}capitalCity\end{myfont} encodes the relation between \begin{myfont}Paris\end{myfont} and \begin{myfont}France\end{myfont} we are looking for. When applied to the second half of the analogy
\begin{myfont}
(capitalCity ?X CityOfLondonEngland)
\end{myfont}
The KB will correctly return \begin{myfont}England\end{myfont} for X.

We can find such predicates using the following query in the Cyc KB:

\begin{myfont}(and (and 

(assertedPredicateArg France 1 ?X) 

(assertedPredicateArg CityOfParisFrance 2 ?X)) 

(?X France CityOfParisFrance))\end{myfont}

There are a few complications with this. First, there may be several possible terms in the KB corresponding to the natural language terms. (For example, there are at least 22 other cities in the world named Paris, not to mention Paris of Troy, Paris Hilton, a half dozen movies entitled \textit{Paris}, etc...) Second, there may be more than one predicate relating such senses to each other: besides being the capital city of France, Paris is also a geographical subregion of France. To deal with these complications, we find all the potential predicates which satisfy \begin{myfont}(predicate France Paris)\end{myfont} for any sense of \begin{myfont}France\end{myfont} and \begin{myfont}Paris\end{myfont} and apply them to all senses of the term \begin{myfont}London\end{myfont}. This returns several potential answers. These answers are then tested to find which is closest to the answer vector obtained using the DSVS to solve the analogy. In this way, we find an answer which is both hand-curated (so more likely to be correct) and most likely in terms of distributional frequency.

Dedre Gentner called such matches with a predefined relation vector relating terms trivial, and felt that very few analogies would be captured by such relationships \cite{Gentner1983}. There are several other ways terms can be related in the knowledge base than those relationships captured by a single predicate.  Another query pattern searches for two concepts and two predicates subject to certain constraints. In the analogy \begin{myfont}taking::took::running:ran\end{myfont}, Cyc only can find the link between \begin{myfont}taking\end{myfont} and \begin{myfont}took\end{myfont} by linking both to the root word \begin{myfont}take\end{myfont}. The predicates relating these terms to \begin{myfont}take\end{myfont} are the same predicates relating \begin{myfont}running\end{myfont} and \begin{myfont}ran\end{myfont} to the root word \begin{myfont}run\end{myfont}, so by searching for a pattern linking pairs of predicates to terms they have in common in this way we can find analogies of this type. Rearranging the roles played by the terms allows the system to also solve the analogy taking:running::took:ran. For these tests we use only a few of the simplest such patterns, as the search becomes more costly and the number of possible answers multiplies for more complex pattern-matching.

When no such pattern can be matched, we can still make use of information in the KB to choose the best answer from among the closest terms to the result vector in the DSVS. One simple way is by checking that the part-of-speech of the answer is correct. In nearly all four-term analogies, the part-of-speech of the fourth term either corresponds to the part-of-speech of the second term (in which case, the first and third terms should also have the same part-of-speech) or corresponds to the third term (in which case the first and second terms should have the same part-of-speech.) A similar relation should hold for singular or plural. \footnote{In practice, our identification of part-of-speech is not perfect, so when no term matching part-of-speech can be found among the near results of the answer vector, we ignore this criterion.}

We tested analogy-finding ability on two test sets: the Semantic Syntactic Word Relationship test set introduced in \cite{Mikolov2013} to test word2vec, and the SAT four-term analogy test set from (Turney, 2005) 

\subsection{Experiment: Semantic Syntactic Word Relationship test set}

For the `DSVS alone' condition, we computed the vector from term 1 to term 2, and added that vector to term 3 to get the answer vector. The natural language term with the minimum distance from this answer vector was used (as long as it was different from the three search terms.)
For the KB alone condition, the KB was searched for results matching the analogy patterns discussed above. When the KB returned many results, an answer was chosen randomly from among the answers returned.

For the combined system, the answer returned by the KB closest to the answer vector of the DSVS was chosen. 

\subsubsection{Results}

\begin{table}[h]
\caption{Percent correct on SSWR test set}
\label{sample-table}
\centering
\begin{tabularx}{0.95\textwidth}{ |l|X|X|X| }
\hline
category &DSVS alone	&KB alone	&Displacer (DSVS + KB)\\\hline
capital-common-countries	&83	&52	&98\\
capital-world	&80	&66	&98\\
currency	&35	&52	&70\\
city-in-state	&72	&38	&92\\
Family	&85	&10	&86\\
adjective-to-adverb	&29	&70	&86\\
opposite	&43	&0	&48\\
comparative	&91	&69	&93\\
superlative	&87	&4	&95\\
participle	&78	&100	&100\\
nationality-adjective	&90	&4	&90\\
past-tense	&66	&77	&78\\
plural	&90	&89	&97\\
plural-verbs	&68	&93	&100\\
\hline
\end{tabularx}
\end{table}

\subsubsection{Discussion}

It should not be surprising that a database of countries and capitals is able to do a good job of finding the capital of a given country. However, Displacer performed better than either the KB alone or the DSVS alone on all categories. The KB was unable to find matching patterns for many analogies in some categories, leading to a low score on opposites, superlatives, nationality, and family. In these cases, Displacer relied on the DSVS, only enforcing correct part-of-speech. For other predicates, the KB found too many potentially correct answers. Without a good way to judge between these for what the most natural answer should be, the system chose randomly between them, and only guessed the ground truth answer a small fraction of the time. This explains the low score for the KB on capitals, currency, and cities-in-states. In this case, Displacer found which of the potential answers returned by the KB was closest to the vector in the DSVS.

The KB has not explicitly been told the correct form for the part of speech for every word in English. Instead, it has rules to produce the form from the word root, and some exceptions, which don't cover every case. Instead of the word \begin{myfont}sang\end{myfont}, for example, the KB returned only the incorrect result \begin{myfont}singed\end{myfont}. Because the KB can be unreliable in this way the system could be improved by flagging results where the KB result differed from the DSVS result. In general, when the same answer is obtained using both the KB and the DSVS it is likely to be correct, while if they are wildly different, the answer is suspect.
\subsection{Experiment: SAT four-term analogy test set}

The second set of 374 four-term analogies comes directly from SAT tests \cite{Turney2005}. Unlike the analogies in the previous test, these are meant to be difficult for a human to solve. In the past, most testing of automated systems on these questions has been done by choosing among multiple-choice answer pairs, but to make the results comparable with the previous test, we provide the first three terms of the analogy and ask it to provide an appropriate fourth term. \footnote{Since these analogies allow for several possible fourth terms, there is a subjective element to the scoring.}

\subsubsection{Results}

\begin{table}[h]
\caption{Percent correct on SAT four-term analogy test set}
\label{sample-table}
\centering
\begin{tabularx}{0.85\textwidth}{ |X|X|X| }
\hline
DSVS alone	&KB alone	&Displacer (DSVS + KB)\\\hline
12 &0 &24\\
\hline
\end{tabularx}
\end{table}

\subsubsection{Discussion}

Most of this gain comes from enforcing part-of-speech constraints and rejecting synonyms and related forms of the three input terms when appropriate to do so. In only ten cases did the attempts to find an appropriate analogy in the KB successfully improve the result over the DSVS alone, and in four cases it caused the system to falsely reject a good answer. Clearly the system's approach to finding analogies in the KB is too simplistic for more difficult analogies.

\section{Conclusion and Future Work}

A KB and DSVS can be used together to make the KB less brittle and with greater coverage (with approximate knowledge), and to provide enough analogous examples to the DSVS to allow it to obtain good estimates of the probability of the correctness of its own answers. On both query answering tasks and analogy tasks, the combination of the two outperforms either working alone.

 In this paper all of the queries are expressed in Cyc's syntax, but to be widely useful a system must be able to handle natural language queries. Displacer itself could potentially be used for some level of word-sense disambiguation and reference resolution. We plan to look at ways to confirm or reject possible answers based on other information within the KB or DSVS that changes the probability of an answer being true. We also plan to explore how learning from image or video sources might be used to augment the information learned from text.

\newpage

%
\nocite{*}
\label{sect:bib}
\bibliographystyle{plain}
\bibliography{semantickb}


\end{document}